\documentclass{article}

\usepackage{arxiv}

\usepackage[utf8]{inputenc} 
\usepackage[T1]{fontenc}   
\usepackage[dvipsnames]{xcolor}
\usepackage[colorlinks=true, linkcolor=RoyalBlue, citecolor=Green, urlcolor=Blue]{hyperref}  
\usepackage{url}           
\usepackage{booktabs}      
\usepackage{amsfonts}      
\usepackage{nicefrac}       
\usepackage{microtype}     
\usepackage{lipsum}		
\usepackage{graphicx}
\usepackage{natbib}
\usepackage{doi}

\title{A Review on Self-Supervised Learning for Time Series Anomaly Detection: Recent Advances and Open Challenges}

\renewcommand{\shorttitle}{A Review on Self-Supervised Learning for Time Series Anomaly Detection}

\author{
  \href{https://orcid.org/0000-0001-6127-0686}{Aitor Sánchez-Ferrera}\\
  University of the Basque Country UPV/EHU \\
  Manuel Lardizabal Ibilbidea 1 \\
  Donostia/San-Sebastián, Spain \\
  \texttt{aitor.sanchezf@ehu.eus} \\
  \And
  \href{https://orcid.org/0000-0001-9969-9664}{Borja Calvo} \\
  University of the Basque Country UPV/EHU \\
  Manuel Lardizabal Ibilbidea 1 \\
  Donostia/San-Sebastián, Spain \\
  \texttt{borja.calvo@ehu.eus} \\
  \And
  \href{https://orcid.org/0000-0002-4683-8111}{Jose A. Lozano} \\
  University of the Basque Country UPV/EHU \\
  Manuel Lardizabal Ibilbidea 1 \\
  Donostia/San-Sebastián, Spain \\
  Basque Center for Applied Mathematics (BCAM) \\
  Mazarredo Zumarkalea 14 \\
  Bilbao, Spain \\
  \texttt{ja.lozano@ehu.eus} \\
}

\renewcommand{\headeright}{ }
\renewcommand{\undertitle}{}
\renewcommand{\shorttitle}{A Review on Self-Supervised Learning for Time Series Anomaly Detection}

\hypersetup{
pdftitle={A template for the arxiv style},
pdfsubject={q-bio.NC, q-bio.QM},
pdfauthor={David S.~Hippocampus, Elias D.~Striatum},
pdfkeywords={First keyword, Second keyword, More},
}

\begin{document}
\maketitle

\begin{abstract}
Time series anomaly detection presents various challenges due to the sequential and dynamic nature of time-dependent data. Traditional unsupervised methods frequently encounter difficulties in generalization, often overfitting to known normal patterns observed during training and struggling to adapt to unseen normality. In response to this limitation, self-supervised techniques for time series have garnered attention as a potential solution to undertake this obstacle and enhance the performance of anomaly detectors. This paper presents a comprehensive review of the recent methods that make use of self-supervised learning for time series anomaly detection. A taxonomy is proposed to categorize these methods based on their primary characteristics, facilitating a clear understanding of their diversity within this field. The information contained in this survey, along with additional details that will be periodically updated, is available on the following GitHub repository: \url{https://github.com/Aitorzan3/Awesome-Self-Supervised-Time-Series-Anomaly-Detection}.
\end{abstract}

\keywords{Anomaly detection \and self-supervised learning \and time series analysis \and outlier detection \and taxonomy}

\section{Introduction}

Time series refer to collections of measurements or recordings arranged in chronological order \cite{hamilton2020time}. The categorization of time series depends on the number of dimensions involved. A time series is classified as univariate if it is based on the evolution of a single variable over time. Conversely, a time series is referred to as multivariate if it is composed of multiple univariate time series that collectively describe a single system or entity. Mining time series data involves various tasks such as classification, regression, clustering, and anomaly detection, which yield valuable insights in domains like economy, health, and industry \cite{cryer1986time, fu2011review, esling2012time}. In recent years, the advent of powerful machine learning techniques has sparked significant interest in time series anomaly detection, enabling effective solutions for tasks like financial fraud detection \cite{awoyemi2017credit, chaudhary2012review}, network traffic monitoring \cite{ahmed2016survey, kwon2019survey}, disease detection \cite{ukil2016iot, bao2019computer}, and fault detection in Internet of Things (IoT) devices \cite{thanigaivelan2016distributed, lin2020anomaly}, among others.

Time series anomalies, also known as novelties or outliers \cite{carreno2020analyzing}, are defined as abnormal events that deviate from the expected behavior in the system under analysis \cite{schmidl2022anomaly}. As mentioned in \cite{aggarwal2017introduction}, these phenomena can be interpreted in two distinct ways. They can either signify abnormal events that we aim to identify, or they can be inaccurate and noisy measurements that we seek to eliminate or rectify in order to enhance the quality of our dataset. For the purpose of this study, we will refer to anomalies as the former. The literature identifies various types of anomalies based on their context \cite{blazquez2021review}:

\begin{itemize}
    \item Local Context Anomalies: these deviations manifest at the level of observations within the context of an individual time series. This type of anomaly detection zooms in localized patterns of abnormality, allowing for a more detailed examination of deviations at the granular level. We identify two primary types of local anomalies:
    \begin{itemize}
        \item Point Outliers: these anomalies pertain to specific timestamps where the values within a time series exhibit a substantial deviation from other values or neighboring points. Essentially, point outliers stand out as isolated data points that significantly differ from their surrounding data, contributing to irregularities in the overall temporal pattern.
        \item Subsequence Outliers: this category of anomalies comprises consecutive points in time that collectively display atypical behavior, forming a subsequence of abnormality within the time series. In other words, subsequence outliers manifest as clusters of consecutive data points exhibiting unusual patterns or trends, contributing to an identifiable deviation from the expected temporal behavior.
    \end{itemize}
    \item Global Context Anomalies: this anomaly category involves entire time series that function as outliers. Specifically, tasks related to anomaly detection in this context revolve around identifying complete time series describing anomalous events within a database that comprises numerous time series. In essence, the focus is on detecting overarching patterns or sequences of abnormal behavior across the entirety of a time series dataset, emphasizing the comprehensive nature of the anomalies being addressed.
\end{itemize}

\begin{figure}
    \centering
    \includegraphics[width=\linewidth]{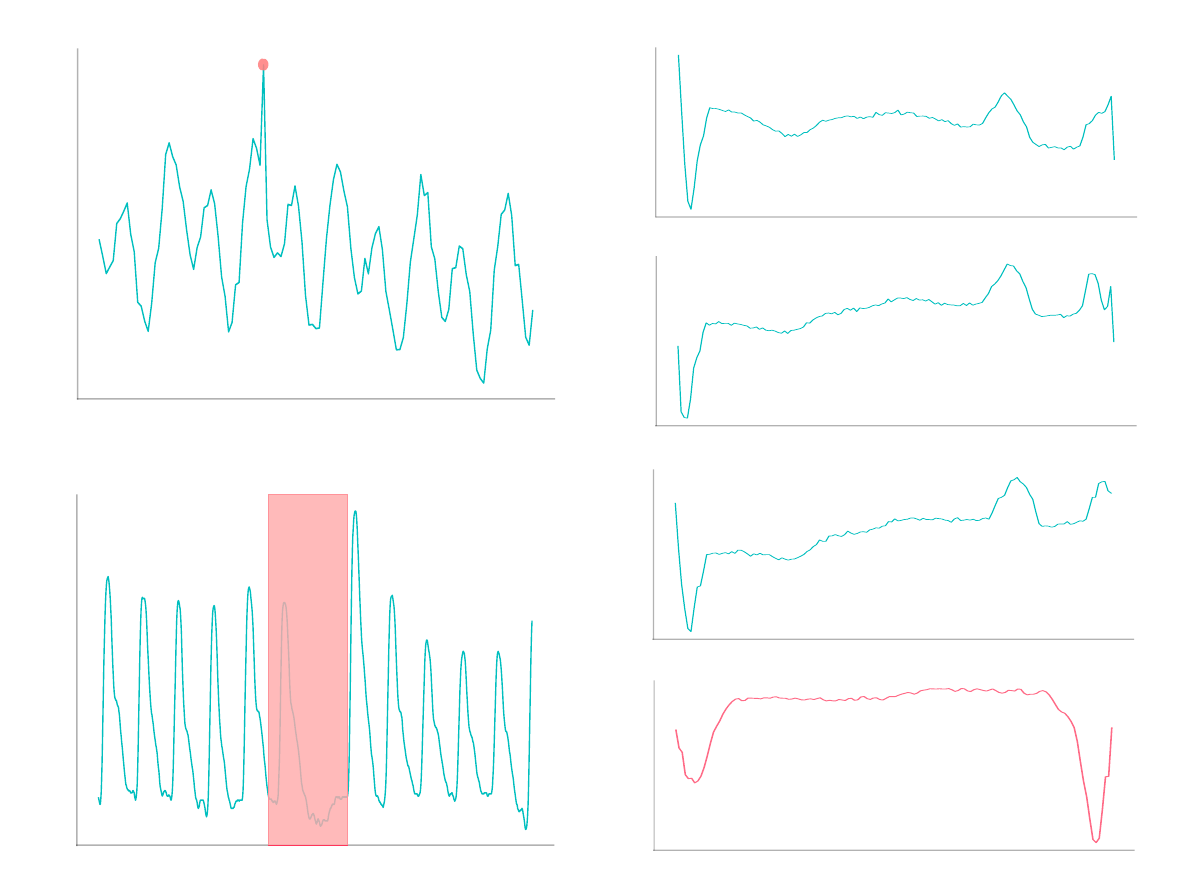}
    \caption{Examples of time series anomalies. Top left: a point anomaly belonging to the \textit{Yahoo-TSA} dataset. Bottom left: a subsequence anomaly belonging to the \textit{UCR\_BIDMC1\_2500\_5400\_5600} dataset. Right: a sequence anomaly in the global context of the \textit{ECG5000} dataset.}
    \label{fig:anomalies}
\end{figure}

Figure \ref{fig:anomalies} depicts real examples of the different types of time series anomalies explained before.
\vspace{4mm}

In the field of machine learning for anomaly detection in time series data, two main perspectives are distinguished: supervised learning-based approaches and unsupervised learning-based approaches \cite{darban2022deep}. In the supervised setting, models are trained using datasets where normal and abnormal samples are labeled. Typically, a supervised predictor in an anomaly detection scenario functions as a binary classifier, learning to predict whether a new sample is an anomaly based on labeled samples. While these models can achieve high accuracy in anomaly detection, training a supervised classifier for anomalies can be problematic. Anomalies, by their nature, are infrequent occurrences, resulting in an extremely imbalanced classification problem that is challenging to solve. Additionally, creating labeled datasets for training requires significant time and specialized knowledge related to the specific problem, making supervised training of machine learning models often unfeasible in real-world contexts \cite{shaukat2021review}.

Unsupervised approaches, unlike supervised methods, do not rely on labeled datasets. Instead, they are trained using datasets composed solely of normal samples, aiming to capture the underlying patterns and characteristics of the data distribution in the problem at hand. Once trained, these approaches typically compare the properties of new samples with the learned notion of normality from the training phase to assess their abnormality degree or anomaly score \cite{pang2021deep}. If the anomaly score of a new sample exceeds a predetermined threshold, it is classified as an outlier. One popular example of an unsupervised method for time series anomaly detection is the use of autoencoders \cite{bank2020autoencoders}. By training an autoencoder exclusively on normal samples, we assume that it learns the latent subspace representing the normal patterns and characteristics of the data. During the inference phase, the reconstruction error of the trained autoencoder serves as the anomaly score for detecting anomalies within new samples.

Unsupervised methods, despite their flexibility and independence from labeled datasets, can indeed suffer from overfitting to the observed normality during the training phase. To mitigate this issue, it is crucial for unsupervised approaches to be trained on datasets that encompass a wide range of diverse normal samples. This allows them to capture the complete spectrum of normal characteristics present in the data distribution. Otherwise, during the inference phase, if unsupervised methods encounter normal samples that exhibit slight deviations from the ones in the training set, they may struggle to generalize and mistakenly classify them as anomalies. This can lead to an increased false-positive rate in anomaly detection. Therefore, the effectiveness of unsupervised methods relies heavily on the quality and representativeness of the normal training data they are exposed to \cite{muruti2018survey}.

New perspectives are emerging as a solution to the limitations of unsupervised methods, with self-supervised learning gaining significant attention. Self-supervised learning is an unsupervised learning paradigm that focuses on representation learning, that is, learning representations of the data that make it easier to extract useful information for building classifiers and predictors \cite{6472238}. Thus, self-supervised learning aims at enhancing the ability of machine learning models for capturing the underlying distribution of data and improve their performance across various tasks. Note that the representations learned by means of this methodology are general and to some extent "agnostic" to the final downstream task(s) that we want our models to accomplish (anomaly detection in our case) \cite{liu2021self}.

Recently, there has been a substantial increase in the number of contributions utilizing self-supervised learning for time series anomaly detection. While existing literature includes reviews on machine learning for anomaly detection \cite{chandola2009anomaly, chalapathy2019deep, pang2021deep}, as well as some surveys focusing on anomaly detection in temporal data \cite{gupta2013outlier} or time series specifically \cite{blazquez2021review, braei2020anomaly, choi2021deep}, in addition to a review on self-supervised learning for time series analysis \cite{zhang2023self}, to the best of our knowledge, no survey has specifically addressed the growing use of self-supervised learning for time series anomaly detection. Furthermore, none of the previous surveys cover the extensive range of recent contributions in this area. Therefore, there is a need for a comprehensive review that categorizes these contributions and aids researchers in identifying future research directions in self-supervised time series anomaly detection. This paper aims to address this gap and presents three key research contributions:

\begin{itemize}
    \item A review of self-supervised learning-based time series anomaly detection approaches: the paper presents a comprehensive review of the contributions that utilize self-supervised learning methods for time series anomaly detection. To the best of our knowledge, this is the first survey that focuses exclusively on the use of self-supervised learning approaches for anomaly detection in time series data, filling a gap in the existing literature.
    \item A taxonomy for self-supervised time series anomaly detection approaches: we propose a taxonomy for self-supervised approaches used in anomaly detection for time series data. This taxonomy is developed based on the key characteristics of self-supervised methods found in the literature, and it enhances the overall understanding of the existing contributions in this field by capturing the properties of each method.
    \item An analysis of research challenges and future directions: the paper examines the current research challenges in the field of self-supervised learning for time series anomaly detection. It also identifies and discusses future research directions to improve the performance of anomaly detection using self-supervised learning in time series data.
    \item A compilation of available software and datasets: we collect the source code of methods associated with the contributions examined in this study, along with the time series datasets utilized in their experiments, and provide information on where they can be accessed.
\end{itemize}

It is worth noting that self-supervised learning is often employed for pre-training models and afterwards they are fine-tuned using labeled data or in conjunction with semi-supervised learning \cite{zhu2023understanding, gui2023survey}. However, as annotated data about anomalies is not commonly available, unsupervised approaches for anomaly detection are the most practical option \cite{hojjati2022self}. Due to this, in this survey, we focus solely on analyzing unsupervised methods that do not rely on the use of labeled anomalies.  

The rest of the article is organized as follows. Section \ref{sec:overview} provides an overview of self-supervised learning. In Section \ref{sec:taxonomy}, we present a taxonomy for self-supervised time series anomaly detection approaches. Sections \ref{sec:local} and \ref{sec:global} explore the works that build upon self-supervised time series anomaly detection and analyze their properties. Section \ref{sec:software} compiles available software relevant to self-supervised learning in this context, along with the most popular datasets used for evaluating these methods. Finally, Section \ref{sec:conclusions} presents a discussion of findings and conclusions, summarizing the contributions of the research and suggesting future avenues of exploration. In the interest of ensuring reproducibility, the methodology employed to conduct this survey is presented in Appendix \ref{sec:methodology}.

\section{An overview of self-supervised learning}\label{sec:overview}

In Yann LeCun's statement at AAAI 2020, self-supervised learning was originally described as a process where `a machine predicts parts of its input based on observed parts' by making predictions about information within the input that is assumed to be unknown. However, self-supervised learning has evolved to encompass a broader set of techniques and objectives, including predicting certain data attributes, relationships, or transformations within the data without relying on human annotations \cite{liu2021self}. This approach falls within the realm of unsupervised learning and leverages the abundance of unlabeled data to capture the underlying structure and patterns of the data distribution, thereby enhancing the learned latent representations of machine learning models  \cite{zhai2019s4l}. Traditional self-supervised pipelines typically consist of two main components: the self-supervised pretext task and the downstream task \cite{jing2020self}.

The self-supervised pretext task, also known as proxy task, involves predicting specific attributes or characteristics of data that are already known. Essentially, the pretext task is automatically created based on certain data attributes, relationships, or transformations within the data without relying on human annotations, leveraging large-scale unlabeled data. In self-supervised learning, a model learns to solve the pretext task by assuming that it will gain a deep understanding of valuable data information and the data's inherent structure, which will result in the model capturing useful representations of data \cite{gui2023survey}. These learned representations can enhance the model's performance when applied to the final downstream task, such as classification, regression, or anomaly detection, which is the ultimate goal of the learning process. The model can learn the pretext task together with the downstream task in a multi-task way, or by performing a pre-training in the proxy task and then fine-tuning the model for the downstream task. Note that, while the pretext task may not directly align with the ultimate downstream task, a well-designed pretext task is expected to effectively guide in learning valuable and versatile data representations that benefit the performance in the final downstream task \cite{misra2020self}.

As mentioned before, the self-supervised proxy task can be introduced in the learning process of the model in two primary manners: i) through pre-training a model on the proxy task or ii) by simultaneous learning it together with the downstream task. In both scenarios, the goal is to leverage the knowledge obtained from the self-supervised pretext task for one (or more) downstream task(s), but the execution of the proxy task and the transfer of this knowledge are achieved in different ways \footnote{The methods outlined below are presented within the perspective of deep learning, as the majority of self-supervised learning techniques are rooted in this field. Nevertheless, note that self-supervised learning can find application in broader contexts that do not exclusively depend on deep learning.}.

The approaches considering self-supervised pre-training involve two primary steps. First, a module $f_{p}$, often referred to as the `proxy module' (typically a neural network with multiple hidden layers), is trained to solve the proxy task. This module learns to map the input space of unlabeled data to the output space of the proxy task, $f_{p} : X \rightarrow Y_{p}$. This initial phase is known as self-supervised pre-training. Once the self-supervised pre-training is complete, we proceed to train the `downstream module' $f_{d}$. This module is trained by leveraging the representations learned by the proxy module to solve the downstream task, a process referred to as fine-tuning. Specifically, in this phase, we use the input data $X$ of the downstream task as input for the proxy module $f_{p}$, extracting its hidden representations from one or more hidden layers and discarding the outputs related to the proxy task. This generates a new representation of the input data $\phi(X)$  that is employed as input for the downstream module, which is trained for the specific downstream task at hand, resulting in $f_{d}: \phi(X) \rightarrow Y_{d}$. It is worth noting that the earlier layers of the proxy module capture more general data patterns, while the later layers tend to capture task-specific attributes \cite{gui2023survey}. The fundamental assumption behind this process is that representations learned by effectively solving a well-designed proxy task hold value and can be subsequently applied to various downstream tasks related to the input data within the same dataset \cite{zhu2023understanding}. 

In the literature, the hidden layers of the proxy module used to extract useful representations for the downstream module are commonly referred to as the `feature extractor' $\phi(\cdot
)$, while the remaining layers of the proxy module and the downstream module are termed the `task-specific heads'. These task-specific heads are responsible for mapping the hidden representations of the input data extracted from the pre-trained feature extractor to the output space corresponding to their respective tasks. It is important to note that in the second step, the parameters of the feature extractor are typically frozen to ensure that the learning of the downstream task does not affect the representations learned during the pre-training phase. However, with the emergence of new techniques in the field of transfer learning, there are contributions that also allow for adjustments to the feature extractor's parameters during the fine-tuning step. This enables the updating of the hidden representations learned during the initial step to make them more specific to the chosen downstream task. 

As an alternative to pre-training, there are other approaches where the pretext task and the downstream task are jointly learned in a multi-task learning manner \cite{zhang2018overview}. In this setup, the downstream task takes precedence as the primary objective, while the proxy task serves as an auxiliary task aimed at enhancing data representations for facilitating the learning of the downstream task. The central assumption here is that well-designed proxy tasks share structural similarities with the downstream task, facilitating the transfer of knowledge from the learned representations of the auxiliary task to enhance the performance of models on the downstream task \cite{ahmed2008training}. In broad terms, both the pretext task and the downstream task are trained using the same foundational or base model, typically a neural network with multiple hidden layers referred to as the feature extractor. Additionally, each task is equipped with its respective task-specific head module ($f_{p}$ and $f_{d}$) and the overall loss of the model is computed as a linear combination of the losses associated with the downstream task and the pretext task. Note that in this scenario, unlike in the second step of the pre-training approach where the parameters of the feature extractor are typically frozen, the learning of the downstream task influences the representations captured by the feature extractor when learning the proxy task. Thus, in this case, the representations learned by the feature extractor will always be tailored to the downstream task at hand, thus not being necessarily good for other downstream tasks.

Overall, in the pre-training scenario the objective is to devise a precise self-supervised proxy task capable of capturing the most valuable representations that enhance the performance of the posterior downstream module(s) (optimally general enough representations that are useful for a wide range of downstream tasks). Conversely, in the second context, the aim is to select auxiliary pretext tasks that enhance the performance of a specific downstream task that we want our model to accomplish. It is worth noting that self-supervised approaches allow for the exploration of multiple proxy tasks to enhance the quality and suitability of acquired representations. Consequently, the incorporation of multiple proxy tasks can significantly improve the overall performance of the model in the respective downstream task(s) \cite{doersch2017multi}.

Self-supervised learning was initially introduced in the field of artificial vision \cite{ahmed2008training, doersch2015unsupervised}. Following its success in learning effective visual representations, the utilization of self-supervised learning was afterwards extended to diverse domains such as medicine and healthcare \cite{krishnan2022self}, graph learning \cite{liu2022graph}, audiovisual data \cite{liu2022audio, schiappa2022self}, recommender systems \cite{yu2023self}, and even remote sensing \cite{wang2022self}. Be aware that the specific design of the proxy task influences the model's ability to capture local patterns, as well as significant properties and dependencies within the data. In addition, it is important to note that proxy tasks in self-supervised learning are often tailored to specific domains. Therefore, one of the major challenges researchers have encountered is devising new pretext tasks that can be applied to other types of data, such as time series data \cite{hojjati2022self}. 

\subsection{Types of self-supervised pretext tasks}\label{sec:typesssl}

Existing literature on self-supervised learning identifies two main types of self-supervised pretext tasks: self-predictive tasks and contrastive learning-based tasks \cite{liu2021self}. Both types of proxy tasks can be considered when applying the self-supervised learning strategies as pre-training and multi-task learning. Recall that the concept of self-supervised learning originated in the artificial vision domain, consequently, the explanations and examples provided in this section will predominantly focus on its application to images. Later, in the literature review, we will focus on time series data.

\subsubsection{Self-predictive pretext tasks}\label{sec:selfpred}

The pretext tasks in the field of self-prediction involve constructing self-supervised prediction tasks at the individual data sample level. Typically, a portion of the data is used to predict another part that is assumed to be unknown. This is often achieved by applying one or more transformations to each sample in the training set for generating a set of augmented views of the samples. The tasks associated with self-predictive self-supervised learning typically revolve around either predicting the specific transformation applied to the original sample to generate each augmented view or reconstructing the original samples based on their augmented views. Broadly speaking, self-predictive tasks can be divided into three primary classes: \textit{self-supervised classification}, \textit{self-supervised reconstruction} and \textit{self-supervised forecasting} \cite{liu2021self}.

\paragraph{Self-supervised classification} The pretext tasks related to this category involve applying a family of transformations $\mathcal{T} = \lbrace T_1, ..., T_K \rbrace$ to each input sample  $x_i$ in the training set. This results in a series of augmented views specific to that sample $\mathcal{T}(x_i) = \lbrace T_1(x_i), ..., T_K(x_i) \rbrace$, where $T_k(x_i)$ denotes the augmented view obtained by applying the $k$-th transformation from $\mathcal{T}$ to $x_i$, and $k$ serves as the pseudo-label associated to $T_k(x_i)$. Then, the self-supervised classification-based pretext task is based on training a classifier $f(\cdot)$ to discriminate among the $K$ transformations applied to each sample for generating the corresponding augmented views, such that ideally $f(T_k(x_i)) = k$ for $1 \leq k \leq K$. This is achieved by minimizing a classification loss, such as cross-entropy. These proxy tasks heavily depend on the adeccuacy of the transformations considered for generating the augmented views of samples. The most popular choice in this field is the use of geometric transformations \cite{golan2018deep, gidaris2018unsupervised}.

\paragraph{Self-supervised reconstruction} The proxy tasks within this category involve applying a transformation $ T $ to each input sample $x_i$ in the training set, resulting in an augmented view of that particular sample $ T ( x_i )$. The pretext tasks within the realm of self-supervised reconstruction are based on training a model $f(\cdot)$ to reconstruct, from the augmented view of each sample in the training set, its corresponding original sample $f(T(x_i)) = x_i$. Thus, the model is trained by minimizing the reconstruction error between the model's output and the original sample. The methods proposed within this field differ primarily on the transformation considered for generating the augmented views and performing the self-supervised reconstruction. Among the prominent choices, we find masking a portion of the data to restore the original sample \cite{pathak2016context}, generating the grayscale version of an image to predict the corresponding color channels of the original image \cite{zhang2016colorful}, or adding noise to images and reconstruct clean data from noisy inputs as it is done with denoising autoencoders \cite{vincent2008extracting}. These tasks, among others, are all oriented to enabling models to grasp the complete content of an image, generating plausible hypotheses for missing information, and understanding the relationship between image semantics and textures.

\paragraph{Self-supervised forecasting} In this category, we find pretext tasks that leverage the temporal dependencies that are inherent in some types of data. Let us consider a time-evolving sample $x \in  \mathbb{R} ^{d \times L}$, where $d$ is the number of dimensions, $L$ is the total number of timestamps, and $x_t \in \mathbb{R}^d$ is the value of the sample at timestamp $t$. The pretext tasks belonging to this category involve training models to, on the basis of a context window $x_{t-w, t} = (x_{t-w}, x_{t-w+1}, ..., x_t)$, predict the future $p$ values $x_{t+1, t+p} = (x_t+1, ..., x_t+p)$ of that sample. The model is trained by minimizing the prediction error between the model's outputs and the ground-truth values of the window to be predicted. As it is intuitive, this type of pretext tasks are only applicable to time-evolving data. In the case of videos, this is applied by predicting future frames on the basis of a series of context frames \cite{schiappa2022self}, and for natural language processing, this is achieved by predicting future sentences \cite{kenton2019bert}.

\subsubsection{Contrastive pretext tasks}\label{sec:contrastive}

Contrastive learning is a self-supervised learning paradigm that aims at training models by emphasizing the similarities and differences (or contrasts) between pairs of data samples or parts of them. The models trained by means of these pretext tasks map the input data into a more compact and abstract latent feature space by means of a network that acts as an encoder. The primary objective is to learn representations in which similar samples are brought closer together in the latent space, while dissimilar samples are pushed apart. The core idea is to encourage the model to learn meaningful representations by maximizing the similarity between positive pairs (similar instances) and minimizing the similarity between negative pairs (dissimilar instances) \cite{jaiswal2020survey}. In this context, the term "similarity" includes metrics that evaluate the resemblance between latent representations of data samples or components, such as cosine similarity, as well as distance metrics like Euclidean or Mahalanobis distances.

Contrastive learning employs an iterative procedure, through which the following steps are repeated for each sample in a batch.
\begin{enumerate}
    \item \textbf{Pair Selection}. For the current sample (the anchor), positive and negative samples are chosen for creating the positive and the negative pairs. This selection can be done according to different criteria, which will be explained later on.
    \item \textbf{Latent Representations}. The latent representations of the anchor, the positive samples and the negative samples are extracted by means of a neural network that acts as an encoder. In the realm of contrastive frameworks, the utilization of `siamese network' architectures is prevalent for this purpose \cite{bromley1993signature}.
    \item \textbf{Contrastive Loss}. The contrastive loss is computed and minimized to facilitate learning the proxy task. Popular contrastive losses include InfoNCE (Noise Contrastive Estimation) \cite{oord2018representation} and NT-Xent (normalized temperature-scaled cross-entropy) \cite{chen2020simple} losses. Notably, these losses involve processing more than one negative sample per iteration of the contrastive task. Another noteworthy loss function in this context is the triplet loss \cite{schroff2015facenet}, which, unlike InfoNCE and NT-Xent, operates with distance measures instead of similarity measures.
\end{enumerate}

By training models using contrastive pretext tasks, the model acquires a deep understanding of the meaningful and informative similarity patterns within the data. This newfound knowledge shows to be highly valuable, as it can enhance the model's performance in a wide range of downstream tasks \cite{jaiswal2020survey}.  It is crucial to emphasize that certain methods exclude the incorporation of negative samples, focusing solely on positive samples for the contrastive proxy task.

The main difference among the existing contrastive tasks in the literature are based on the way they select the positive and negative pairs for each sample. Even if there exist many, there are two main strategies for doing so and, thus, we distinguish two principal methodologies for contrastive tasks: \textit{augmentation contrast} and \textit{sampling contrast}.

\paragraph{Augmentation contrast} This is the most popular strategy for generating pairs in contrastive learning. Augmentation transformations are used at some stage of the process to either generate positive samples, negative samples or both. Commonly, an augmentation transformation $T_{p}$ is applied to the anchor $x_i$, generating an augmented view $x_p$ that serves as its positive sample. The pretext task aims to maximize the similarity between the latent representations of the anchor $x_i$ and the positive sample $x_p$, while minimizing the similarity between the anchor and negative samples $x_n$. These negative samples are obtained either by augmenting the anchor with a transformation $T_{n}$ or by selecting other samples from the batch. Typically, authors choose positive transformations with the goal of enforcing model invariance to perturbations associated with the specific augmentation. For example, a common positive transformation involves introducing Gaussian noise to samples, fostering invariance in the model to noisy samples, as noise often does not alter the class or nature of the samples in many contexts (depending on the problem). Conversely, negative transformations, when employed, aim to disrupt the inherent nature of the data, creating "corrupted" views of the anchor that are presumed to exhibit distinct properties and characteristics compared to the original sample \cite{jaiswal2020survey}. Common negative transformations in contrastive learning include masking, resizing, color distortion, rotation, and the application of filters \cite{chen2020simple}. However, the selection of these transformations for generating positive and negative samples depends on the context of the problem and the assumptions made by the authors about the data \cite{cai2020all}. It is noteworthy that some studies combine multiple transformations to augment samples and generate pairs in contrastive tasks \cite{chen2020simple}.

\paragraph{Sampling contrast} This approach avoids dependence on transformations for the selection of pairs in contrastive proxy tasks. Instead, it relies on leveraging specialized knowledge about the data to make assumptions regarding the similarity of various instances or components of the data to select positive and negative samples. For instance, in enhancing image classification, positive samples for the anchor can be chosen based on whether they belong to the same class in the classification problem \cite{yang2023label}. In the context of video representation learning, an assumption can be made that consecutive frames of a video should exhibit a similar representation in the latent space. Consequently, contrastive pairs for the anchor can be generated by considering distances between frames in a video \cite{mobahi2009deep}.

The techniques outlined in this section revolve around different types of methods within self-supervised learning. It is evident that the choice of pretext tasks in self-supervised learning must align with the specific data type, rendering it unfeasible to employ all proxy tasks universally for all data types. In the following sections, we focus on the established methods that employ self-supervised proxy tasks for handling time series data and do not rely on the use of labeled data, considering anomaly detection as the downstream task at hand.

\section{A taxonomy for Self-supervised learning-based anomaly detection in time series}\label{sec:taxonomy}

Self-supervised techniques for identifying anomalies in time series data can be categorized based on two primary axes derived from the content of the previous sections. First, the focus of the time series anomaly detector concerning the context in which anomalies are detected. Second, the type of pretext tasks considered during model training.

\begin{figure}
    \centering
    \includegraphics[width=0.7\linewidth]{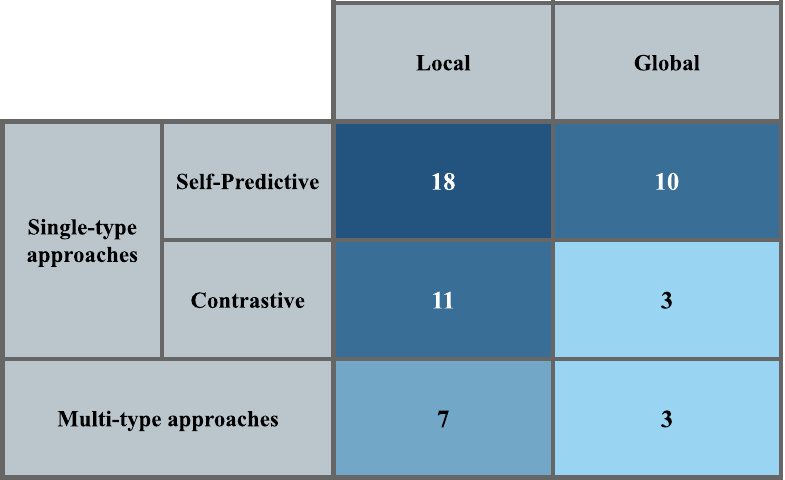}
    \caption{Taxonomy for self-supervised time series anomaly detection approaches.}
    \label{fig:taxonomy}
\end{figure}

\subsection{Context of the anomaly detection task}

As noted in the introduction, time series anomaly detection tasks are commonly categorized into two main scenarios. This axis distinguishes approaches that address the time series anomaly detection task in the following two contexts:

\begin{itemize}
    \item \textbf{Anomaly detection in the local context}. Anomaly detection in the local context refers to tasks focused on identifying unusual patterns that differ from the typical patterns and characteristics within an individual time series. Anomalies within this context may consist of individual points that are abnormal or subsequences of the time series that exhibit abnormal behavior.
    \item \textbf{Anomaly detection in the global context}. This type of time series anomaly detection involves identifying abnormalities within a dataset containing multiple complete time series. Here, the anomalies being detected are instances of entire time series that display anomalous behavior compared to the normal properties exhibited by the rest of the samples in the dataset.
\end{itemize}

It is worth noting that certain studies shift the anomaly detection task from the global context to the local context. One such approach involves dividing an individual time series (local context) into various subsequences and employing a method that treats each resulting subsequence as a sample within a dataset containing all the subsequences (global context). In this survey, we will classify each method based on this axis according to the problem they aim to address. For instance, techniques following the aforementioned methodology will be classified as anomaly detectors in the local context, as they aim to identify abnormal subsequences.

\subsection{Type of pretext tasks}

As mentioned before, there exist two types of pretext tasks, namely self-predictive and contrastive. Furthermore, multiple self-supervised pretext tasks can be combined during the training of the models for better capturing the normality of data. There are some approaches that consider only a single type of proxy task, while others combine both self-predictive and contrastive pretext tasks in the training of the models. Based on this, we can distinguish between two primary categories of approaches:

\begin{itemize}
    \item \textbf{Single-type approaches}. These approaches exclusively rely on a single type of pretext task. Specifically, any approach that exclusively employs a single pretext task falls within this category. Moreover, within the set of approaches that incorporate multiple pretext tasks, those that utilize only one type of pretext task also fall into this same group. In this category, we discern methods that rely exclusively on \textbf{self-predictive} pretext tasks and those that exclusively focus on \textbf{contrastive} pretext tasks.
    
    \item \textbf{Multi-type approaches}. In this scenario, both self-predictive and contrastive learning objectives are integrated in the training of anomaly detection models. It is important to note that, unlike the single-type approaches, where models can be trained using one or more pretext tasks, in this case, models always incorporate a minimum of two proxy tasks in their training.
\end{itemize}

Considering these two fundamental dimensions and the various values they encompass, we can delineate six distinct categories of contributions. The taxonomy established for self-supervised time series anomaly detection methods is visually depicted in Figure \ref{fig:taxonomy}, together with the number of analyzed contributions belonging to each category. In the subsequent sections, we will expound upon each of these potential categories. Specifically, the following two sections are structured based on the first axis of the taxonomy, addressing the context of the anomaly detection task. Within these sections, we further partition the subsections according to the remaining axis of the taxonomy, which relates to the types of pretext tasks considered in the approach. 

\section{Anomaly Detection Methods in the Local Context}\label{sec:local}

Most of the self-supervised methods for time series anomaly detection focus on identifying anomalous patterns within the local context of individual time series $X = \lbrace x_1, x_2, ..., x_L \rbrace \in \mathbb{R} ^ {d \times L}$, where $d$ denotes the number of variables considered ($d=1$ for univariate time series) and $L$ represents the sequence length. The primary objective of these methods is to detect anomalous points $x_t \in \mathbb{R}^d$, anomalous subsequences $X_{t,w} = \lbrace x_{t-w+1}, ..., x_t \rbrace \in \mathbb{R}^{d \times w}$, or both, within the target time series $X$. 

To achieve this goal, these models typically learn one or more proxy tasks tailored to capture the normality of data based on a designated normal region of $X$, which serves as the training set. Typically, time series are segmented by using sliding windows and the proxy tasks are computed based on the resulting subsequences. Subsequently, the knowledge acquired through these proxy tasks is utilized to determine the anomaly score associated with new points and/or subsequences within $X$, constituting the evaluation set.

The studies falling into this category, along with their specific properties, are illustrated in Table \ref{tab:local}. It specifically outlines, the method employed in the self-supervised task(s) considered, the type of outliers that the method identifies, the measure considered for computing anomaly scores, and whether the approach addresses univariate or multivariate time series. In relation to the anomaly scores column, it delineates various approaches for its computation: evaluating the model's classification performance (encompassing classification losses and misclassifications), assessing residual errors (covering reconstruction and prediction errors), and utilizing metrics associated with the latent representations produced by the model (involving representation similarities and distances). We include the use of contrastive losses within the last subgroup, as these compute similarity and distance measures between the latent representations of data parts or samples.

\begin{table}
\begin{center}
  \caption{Summary of self-supervised methods for time series anomaly detection in the local context}
  \label{tab:freq}

  \begin{tabular}{lccccc}
    \toprule
    Paper & Type of PT & Anomaly Type &Anomaly Score & \multicolumn{2}{c}{Dim}\\
    \cline {5-6} & & & &U & M \\
    \midrule
    Self-predictive & & & & \\
    \hline
    \cite{meng2019spacecraft}  & Rec & Point&  RE & &\checkmark\\
    \cite{chen2020anomaly} & Rec & Point &  RE & &\checkmark\\
    \cite{sakurada2014anomaly}  & Rec &Point &  RE & &\checkmark\\
    \cite{audibert2020usad} &  Rec &Point&  RE & \checkmark&\\
    \cite{zhang2019auto} &  Rec &Subseq&  RE & \checkmark&\\
    \cite{marchi2015novel} &  Rec &Point, Subseq &  RE & \checkmark&\\
    \cite{jiang2017wind} & Rec &Point, Subseq &  RE & &\checkmark\\
    \cite{fu2022mad} & Rec &Point, Subseq&   RE  & & \checkmark\\
    \cite{liu2021deepfib}  & Rec &Point, Subseq &   RE & & \checkmark\\
    \cite{pan2022duma} & For & Point  & RE & & \checkmark\\
    \cite{li2022anomaly} & For & Point  & RE & & \checkmark\\
    \cite{shen2020timeseries} & For & Point  & RE & & \checkmark\\
    \cite{munir2018deepant}& For & Point, Subseq  & RE & \checkmark & \\
    \cite{jeong2023anomalybert}   & Class &Point &   CP& &\checkmark\\
    \cite{huang2022efficient} & Class &Subseq &   CP & &\checkmark\\
    \cite{tran2022self} & Class &Subseq & CP &  &\checkmark\\
    \cite{miao2022unsupervised}  & Rec + For &Point &   RE & &\checkmark\\
    \cite{qi2022mad} & Rec + Class & Point & RE & & \checkmark \\

    \hline
    Contrastive& & & & &  \\
    \hline
    \cite{pranavan2022contrastive}   & SC & Point & RM & & \checkmark\\
    \cite{kang2023tictok}  &  SC &Point & RM & &\checkmark\\
    \cite{yang2023dcdetector}   & SC &Point & RM & &\checkmark\\

    \cite{chambaret2022stochastic}   & AC & Point &  RM & &\checkmark\\
    \cite{carmona2021neural}  & AC & Point &  RM & &\checkmark\\
    \cite{li2023contrastive}   & AC &Subseq &  RM& &\checkmark\\
    \cite{yue2022TS2Vec}  & AC & Point &  RM & & \checkmark\\
    \cite{nguyen2023learning}   & AC &Point &  RM& &\checkmark\\

    \cite{chen2023time} & AC & Point & RM & & \checkmark\\
    \cite{schneider2022detecting}   & SC + AC &Point & RM & &\checkmark\\

    \cite{zheng2023unsafe}  &  SC + AC &Point & RM & &\checkmark\\

    \hline
    Multi-type&  & & & & \\
    \hline
    \cite{zhang2022time} & Rec + SC &Point & RE + RM & & \checkmark \\
    \cite{yang2022unsupervised} & Rec + AC &Point &  RE & & \checkmark \\
    \cite{abilasha2022warping} & Rec + AC & Point, Subseq & RM & \checkmark &  \\
    \cite{zhou2022contrastive} & Rec + SC + AC & Point & RE & & \checkmark\\
    \cite{zhang2022timegrid} & For + SC &Point &  RE & & \checkmark\\
    \cite{xu2022calibrated} & Class + SC & Subseq &  RM & & \checkmark \\
    \cite{choi2023multi}  & Class + SC + AC & Point &  RM & &\checkmark\\

  \bottomrule
  
\end{tabular}\label{tab:local}
\begin{minipage}{13.45cm}
\small  U: Univariate; M: Multivariate // Rec: Reconstruction; For: Forecasting; Class: Classification; SC: Sampling Contrast; AC: Augmentation Contrast // Subseq: Subsequence // CP: Classification Performance; RE: Residual Errors; RM: Representation Measures
\end{minipage}
\end{center}
\end{table}

\subsection{Single-type self-supervised approaches}

Single-type self-supervised learning approaches make use of one or more pretext task, but always of the same type, either self-predictive or contrastive, to detect local anomalies within a target time series.

\subsubsection{Self-predictive approaches}

Methods in this category propose self-predictive proxy tasks to capture the normality of individual time series. Most methods in this group utilize self-supervised reconstruction as a proxy task. This involves applying a transformation to the samples in the training set to generate augmented views for each sample. Subsequently, the proxy tasks revolve around attempting to reconstruct the original samples from their corresponding augmented views.  The augmented views of the training samples can be viewed as 'corrupted' versions of the original input data. By learning to reconstruct the original samples from their augmentations, the model captures the underlying patterns and structures of normal data during the training phase, identifying the expected features and representations of the 'non-corrupted' data. The primary difference among these approaches lies in the type of transformation considered for the self-supervised reconstruction-based proxy task.

Drawing from the concept of 'masked language models' as introduced in the works of \cite{kenton2019bert, lan2019albert}, one of the prevailing strategies involves employing masking as a transformation for self-supervised reconstruction \cite{meng2019spacecraft, liu2021deepfib, fu2022mad, jiang2017wind, chen2020anomaly}. These studies propose a pretext task centered around 'masked time series modeling' to capture meaningful data representations. The authors utilize a random masking technique to replace a portion of the normal input subsequences with either random values or predefined constants. Subsequently, the models are trained to reconstruct the original non-masked subsequences using their masked versions as inputs, thereby minimizing reconstruction errors. Once a self-supervised reconstruction-based model is trained, it is assumed that if the model encounters difficulty in faithfully reconstructing a portion of the data, it suggests the presence of data that diverges from the established patterns in the learned data. This divergence can strongly indicate anomalies or irregularities in the input. Consequently, during inference, the reconstruction errors of points and subsequences are utilized as anomaly scores to categorize them as normals or abnormals.

The works by \cite{meng2019spacecraft, jiang2017wind, chen2020anomaly} conduct inference by inputting non-masked subsequences into their trained models. They then utilize the reconstruction errors associated with specific timestamps to assess the anomaly score of points in the time series. It is worth noting that in the case of \cite{jiang2017wind, chen2020anomaly}, their methods are adapted to perform point outlier detection in time series streaming data, which is updated in real time in an online fashion. Other works, such as \cite{liu2021deepfib, fu2022mad}, reproduce the proxy task during inference by masking parts of the subsequences to be evaluated and reconstructing them using their models. Subsequently, they compute the anomaly scores of both points and subsequences based on their reconstruction errors.

In addition to masking, some studies have explored the use of noise injection as a transformation for self-supervised reconstruction. For instance, the works of \cite{marchi2015novel, sakurada2014anomaly} introduce denoising autoencoders for time series anomaly detection. These models inject Gaussian noise to corrupt input samples and aim to reconstruct their original, non-corrupted versions. During the inference phase, the anomaly score for new timestamps and subsequences is computed based on the model's reconstruction errors.

To end with self-supervised reconstruction, the use of regular autoencoders has been widely extended to the identification of point anomalies \cite{audibert2020usad} and subsequence anomalies in time series \cite{zhang2019auto}. Autoencoders encode time series data into a more compact latent space and then reconstruct the original data on the basis of its latent representations. By considering that the encoding process removes a part of the information of the original data by using a neural network-based transformations (the encoder), these methods are included in the category of self-supervised reconstruction.

Within this category, we also encompass the techniques associated with self-supervised forecasting. In this case, the temporal patterns inherent in time series data are leveraged to construct the self-supervised pretext task. In these works, models are trained to predict future timestamps (next-step prediction) and future subsequences (multi-step ahead prediction) with respect to the current window \cite{li2022anomaly, shen2020timeseries, munir2018deepant}. Then, during the inference, the anomaly score of new points and subsequences are computed by assessing the prediction errors of the models. As a novelty, \cite{pan2022duma} introduces a pretext task centered around next-step prediction using partially masked windows as inputs. The authors argue that, by masking a part of the context window, the representations learned by the models are more robust to noise and capture better the normality of the data.

Alternative approaches in anomaly detection focus on employing self-supervised classification to identify anomalous points and subsequences within individual time series. The approaches within this category primarily differ in the choice of transformations considered for the self-supervised classification.

The works by \cite{huang2022efficient, tran2022self} employ self-supervised classification for identifying anomalous subsequences. In \cite{huang2022efficient}, downsampling is utilized at various rates, with each downsampling rate representing a distinct class for the self-supervised classification task. Conversely, \cite{tran2022self} proposes employing jittering (adding Gaussian noise) and vertically flipping the subsequences to construct a binary self-supervised classification task. This binary classification aims to distinguish between the two resulting augmented views of each original sample. During training, the model is trained to differentiate between these augmented views by minimizing the mean cross-entropy value across the augmented views for each sample. For the detection stage, both approaches replicate the self-supervised classification task and utilize the classification loss as the anomaly score for new subsequences.

The approach presented in \cite{jeong2023anomalybert} adapts self-supervised classification to identify abnormal timestamps in the local context of a time series. To construct the proxy task, the authors propose four different transformations, known as degradations, which involve introducing noise and randomly substituting some values within a time series in different ways. During training, they degrade some segments within the time series to simulate abnormalities. The model is then trained to distinguish between the normal and the degraded timestamps in the time series, that is, performing a binary classification. In the inference phase, no degradation is applied to the new samples, and the model is utilized to classify if each timestamp of the input is anomalous or not.

So far, we have explored the three types of self-predictive pretext tasks for detecting anomalies in local time series data. However, some approaches combine different types of tasks to train their models. After model training, one or more of these proxy tasks are replicated during inference, and the anomaly scores for new segments of the time series data are derived from the model's performance in reproducing these tasks.

For example, \cite{miao2022unsupervised} combines masked reconstruction and masked forecasting to extract more generalized features for pinpointing anomalies at specific timestamps within time series data. They simultaneously train both pretext tasks in a multi-task manner, calculating the overall loss as the sum of the reconstruction errors and the prediction errors of the model. During inference, they repeat this process, determining the anomaly score as a weighted sum of these errors, with the weights adjusted by a fixed hyperparameter.

Finally, in the study conducted by \cite{qi2022mad}, the proposed method involves learning a regular reconstruction alongside a self-supervised classification task. To better capture spatial correlations across various dimensions in multivariate time series data, the authors advocate for the utilization of graph convolutional networks. The model is designed to encode and reconstruct the original subsequences while simultaneously learning the following self-supervised classification task. Each subsequence is augmented once through random permutations, and the model is trained to distinguish between the original subsequence and its corrupted augmented view. During the inference phase, the anomaly score for new timestamps is determined solely by evaluating the reconstruction errors generated by the model. It is important to note that during this phase, the classification task is disregarded and not reproduced for computing the anomaly score of new timestamps.

This last method differs from the approach presented by \cite{miao2022unsupervised} where all proxy tasks are reproduced during inference. Here, the self-supervised classification is introduced during the training phase of the models to enhance the learning of better representations of data in addition to the reconstruction task. Nevertheless, it is the reconstruction task that plays an active role in inferring the anomaly score for new timestamps. This strategy, which involves integrating auxiliary proxy tasks that facilitate the learning of enhanced representations, is employed by several works discussed in the following sections of this paper. These auxiliary tasks are not considered during the inference phase of the models but contribute to the overall improvement of the learned representations.

\subsubsection{Contrastive approaches}

The works in this subsection utilize contrastive proxy tasks to capture the local dynamics of individual time series and conduct anomaly detection to identify unusual points and subsequences. Commonly, time series are preprocessed by splitting them into time windows, and the pretext tasks are computed based on the resulting subsequences. We primarily distinguish between approaches employing sampling contrast and those considering augmentation contrast instead.

We find a few works that build upon sampling contrast, where no transformations are considered for the contrastive proxy task. Instead, authors make various assumptions about data similarities and disimilarities to generate the contrastive pairs. The first proposal we find in this field is \cite{yang2023dcdetector}. Two distinct representations are derived from the input subsequences through self-attention mechanisms: the patch-wise representation, capturing the relationships among points at the same position in each segment, and the in-patch representations, depicting the relations between sample points within the same segment. The contrastive objective aims to bring these two representations together in a shared latent space. Finally, the anomaly score for new timestamps is calculated as the discrepancy between the in-patch and patch-wise representations of the testing data.

The remaining studies employing sampling contrast adopt a methodology known as Contrastive Predictive Coding (CPC), which falls under a subcategory termed `prediction contrast' in the literature \cite{zhang2023self}. CPC focuses on training an encoder to produce representations of subsequences useful for predicting future values. In the works by \cite{pranavan2022contrastive, kang2023tictok}, an encoder, an autoregressive model, and a non-linear projection head are utilized. Initially, the encoder generates the latent representation of each input subsequence. Subsequently, the autoregressive model predicts a context vector for a prediction window extending one or more steps into the future from the current window. Finally, the non-linear projection head estimates the latent representations of the prediction window, and the infoNCE contrastive loss is employed for model training. Essentially, this approach resembles self-supervised forecasting, but with a focus on estimating the latent representation of the future window using a contrastive learning objective, rather than explicitly predicting its content. To simplify, the prediction window is treated as a positive sample to the anchor, while other windows serve as negative samples.

After model training, \cite{pranavan2022contrastive} represents normal sample latent representations in the latent space using a Gaussian distribution. For each new subsequence, they calculate the probability density of its representation with respect to the Gaussian distribution to identify abnormal points. On the other hand, \cite{kang2023tictok} determines abnormal points in time series by modeling the anomaly score through a variant of the CPC loss.

Apart from the previous approaches, there are various studies that build upon using augmentation contrast. In augmentation contrast, positive transformations are typically employed to encourage models to capture invariance with respect to those transformations, assuming they do not alter the normal nature of the data. Conversely, negative transformations, when employed, aim to enable models to detect patterns associated with those transformations, assumed to be indicative of unusual behavior. The primary distinction among works in this field lies in the selection of transformations considered to generate the contrastive pairs.

The work by \cite{li2023contrastive} employs Gaussian noise injection into time windows as a positive transformation, while generating negatives through shuffling, scaling, and changing trends. The proposal of \cite{chambaret2022stochastic}, combines various transformations like noise addition, signal magnitude adjustment, and time warping to create positive samples. However, in this case other windows in the batch are used as negatives. In another scope, \cite{carmona2021neural} generates negative samples by introducing point anomalies in time windows. Unlike the previous approaches, they do not employ any transformation to generate positive samples but assume that neighboring windows should share similar representations in the latent space, designating them as positive pairs. They all utilize similarity measures, such as the contrastive loss or distances between latent representations to compute anomaly scores for new points \cite{chambaret2022stochastic, carmona2021neural} and subsequences \cite{li2023contrastive}.

Another popular method in this category is TS2Vec \cite{yue2022TS2Vec}. TS2Vec is a contrastive framework for time series representation learning based on the following. Each window is randomly masked and cropped twice to generate two augmented views from it. Then, two proxy tasks are learned on the basis of the overlapped regions of the augmented views:

\begin{itemize}
    \item Temporal Contrasting: it brings together the representations of the same timestamp from the two augmented views. At the same time, it pushes away the representations of different timestamps within the same augmented view.
    \item Instance-wise contrasting: it maximizes the similarity between representations of different timestamps within the same augmented view. Simultaneously, it minimizes the similarity with respect to the representation of the same timestamp in other time series in the batch. 
\end{itemize}

Both proxy tasks are concurrently learned to enable the model to effectively capture both low-level and high-level features. In the inference phase, TS2Vec targets the identification of anomalous timestamps in time series. To achieve this, the new time series is split into segments, and the latent representation of each segment is computed twice: initially with the last observation (the timestamp under evaluation) masked, and then without any mask applied. The anomaly score for that point is determined by the distance between these representations. It is important to highlight that the anomaly score of new timestamps is computed in such a way that TS2Vec can be employed for anomaly detection in streaming scenarios. In the work presented by \cite{nguyen2023learning}, a closely aligned contrastive approach is put forth. What sets it apart from the prior contribution is the introduction of noise into overlapping windows instead of masking specific portions. As the earlier method, they train their model employing both the temporal contrastive and the instance-wise contrastive losses. The anomaly score computation remains the same as in TS2Vec, serving as the basis for detecting abnormal timestamps. Notably, this method for calculating anomaly scores for timestamps in time series is adopted by numerous works, which will be further reviewed in this survey.

The previous works have typically relied on manually defined transformations for constructing contrastive pairs to facilitate learning proxy tasks. However, alternative approaches have emerged, advocating for automatic neural transformations that do not rely on assumptions about the most suitable transformations for the data's nature. Studies as \cite{chen2023time, schneider2022detecting, zheng2023unsafe} adopt this strategy, which originated from the proposal of NeuTraL AD \cite{qiu2021neural} (presented in the subsequent section as a method for global time series anomaly detection).

A notable aspect of this line of research is its independence from manually defined augmentations for time series. Instead, these studies leverage neural networks to generate augmented views, which are trained concurrently with other modules in the model. For instance, in \cite{chen2023time}, each subsequence is divided into two overlapping windows: the context window and the suspect window. The suspect window undergoes augmentation through trainable neural transformations. The authors then employ the 'Deterministic Contrastive Loss,' as introduced in NeuTraL AD, to converge the representations of the augmented views and the original input while simultaneously pushing away the embeddings of the augmented views among them. Additionally, they bring together the representations of the context window and the augmentations of the suspect window. During inference, anomaly scores for new timestamps are inferred from the training loss.

The two remaining contributions encompass a fusion of sampling and augmentation contrast methodologies in their proposals \cite{schneider2022detecting, zheng2023unsafe}. In the study by \cite{schneider2022detecting}, they amalgamate the Deterministic Contrastive Loss of NeuTraL AD with Contrastive Predictive Coding during model training. Post-training, the model evaluates the anomaly score of new timestamps utilizing the Dynamic Deterministic Contrastive Loss (DDCL) inherited from NeuTraL AD. In a similar vein, \cite{zheng2023unsafe} employ neural transformations to enhance time series data for contrastive tasks. Their approach involves generating augmented views for each subsequence using neural networks as the positive sample for the augmentation contrast-based task. They ensure that representations of other windows in the batch are pushed apart. Additionally, they concurrently learn a task based on sampling contrast, which entails subdividing the augmented view of the anchor into various subsequences to promote similarity between closely located subsequences and reduce it for those that are more distant. Post-training, they employ the same strategy as TS2Vec for computing the anomaly score of new data points.

\subsection{Multi-type approaches}

This category encompasses techniques employing self-predictive and contrastive proxy tasks for detecting anomalies within a local context. These approaches leverage pretext tasks akin to those discussed earlier. Their primary distinction lies in the choice and combination of pretext tasks used to extract pertinent patterns and features from individual time series.

Most of the studies in this category integrate various proxy tasks with traditional reconstruction using autoencoders. For instance, \cite{zhang2022time} incorporates Contrastive Predictive Coding alongside autoencoder-based reconstruction. During training, both tasks are simultaneously learned, and during inference, the same process is reiterated. The anomaly score for new timestamps is computed by combining the reconstruction error with a modified version of the Contrastive Predictive Coding loss function. 

In the study conducted by \cite{zhang2022timegrid}, focusing on point anomaly detection within sensor networks, models are simultaneously trained on a regular reconstruction-based task, a self-supervised forecasting task, and a sampling contrast-based task. The self-predictive task focuses on next-step prediction relying on the current sliding window. Given the nature of the problem, the contrastive task aims at bringing together the representations of signals from the same sensor while pushing away the representations of signals from other sensors. During training, the model combines the losses from the three tasks to facilitate simultaneous learning. In the inference phase, reconstruction errors are used as the anomaly score for new timestamps.

The proposal by \cite{abilasha2022warping} presents an approach for detecting abnormal points and subsequences. Their method operates under the assumption that data representations should remain invariant to warping-based perturbations for effective anomaly detection. To achieve this, they employ two twin networks (similar to siamese networks but with potentially different weights to learn two self-supervised reconstruction tasks, and an augmentation contrast-based proxy task. Each input subsequence undergoes augmentation through transformations related to time series warping. One of the reconstruction tasks is based on a regular autoencoder, while the other reconstructs the original input using its augmented view. The augmentation-based contrastive task forms contrastive pairs using the anchor and its augmented view as the positive pair. These tasks are simultaneously learned in a multi-task manner during network training. In the inference phase, the anomaly score for new subsequences and points is calculated as the distance between their embeddings and their K-Nearest Neighbors.

In the case of \cite{zhou2022contrastive}, they combine regular reconstruction with a sampling contrast and an augmentation contrast task. The first contrastive task employs contextual contrasting, bringing together the latent representations of adjacent timestamps. In the second task, Gaussian noise is added to the original windows and their frequency-based spectrums to create two augmented views from each subsequence that constitute the positive pair. The three tasks are learned concurrently.  All three tasks are trained simultaneously. During inference, point abnormalities are determined primarily by the model's reconstruction errors, with the contrastive tasks excluded from the anomaly detection stage.

The last contribution that combines reconstruction with other pretext tasks is an augmentation contrast-based pre-training framework \cite{yang2022unsupervised}. The method operates on multivariate time series data, where a variable within the series is randomly selected (the anchor) and augmented through random masking to create a positive sample. Another variable is then chosen as the negative sample. By computing a latent representation that captures both spectral and temporal information, the framework aims to align the positive sample's representation with the anchor while distancing the negative sample's representation. Following pre-training, a decoder module is added to the model, and fine-tuning is conducted based on regular reconstruction. During inference, anomalous timestamps are detected by evaluating the reconstruction errors of new data points.

Other works in this category do not consider the use of regular reconstruction for multi-type self-supervised learning \cite{choi2023multi,xu2022calibrated}. \cite{choi2023multi} builds upon the TS2Vec approach, incorporating pretext tasks from that work along with two additional tasks: temporal consistency and transformation consistency. Temporal consistency is a self-supervised classification task. Given an anchor window, a neighboring window, and a non-neighboring window, the task aims to differentiate between the neighboring and non-neighboring windows. Transformation consistency augments anchor windows using two types of augmentations: weak and strong. Weak augmentation introduces slight variations, while strong augmentation produces substantially different views. This task considers the augmentations of the same anchor as positive samples to the anchor, while treating the weak augmentations of other samples as negatives. The model concurrently learns these tasks, and anomaly scores for new timestamps are computed following TS2Vec's methodology.

Finally, another method belonging to this category is COUTA \cite{xu2022calibrated}. This approach combines self-supervised classification with Deep Support Vector Data (SVDD), a widely recognized sampling contrast method for anomaly detection \cite{ruff2018deep}. SVDD is a deep one-class classification technique that maps normal training data into a higher-dimensional latent space to effectively delineate a hyperplane separating normal instances from potential anomalies. The model is trained to position normal instances close to the center of the hyperspherical latent space, where the center is presumed to embody the normal behavior of the data. Subsequently, outlier detection relies on measuring the distance of a data sample from the hypersphere's center to compute its anomaly score. In COUTA \cite{xu2022calibrated}, an encoder is employed to map time series subsequences to a hyperspherical latent space, aiming to minimize their distance from the center. Unlike in SVDD, the distances of normal representations from the center are modeled as a Gaussian distribution. This addition imposes a penalty on predictions associated with high model uncertainty, effectively addressing issues related to anomaly contamination in the training set. Simultaneously, the model learns the self-supervised classification task, where the transformations considered aim to simulate point and subsequence anomalies (e.g., replacing timestamps with extreme values or substituting segments by random subsequences). Hence, a binary classifier learns to distinguish between normal inputs and their abnormal augmented views. In the inference phase, the model encodes new time series subsequences, and the distance between their latent representations and the center of the hyperspherical latent space is used as their anomaly score.

\section{Anomaly Detection Methods in the Global Context}\label{sec:global}

Self-supervised anomaly detection approaches in the global context aim at identifying abnormal complete time series that deviate from the expected normal patterns and characteristics of a collection of time series samples. Models are trained on a dataset consisting solely of normal time series $\mathbf{X} = \lbrace X_i \rbrace ^N_{i=1}$, where $N$ is the total number of sequences in the dataset, and $X_i \in \mathbb{R}^{d \times L}$ with $d$ variables or dimensions ($d=1$ for univariate time series) and length $L$. The objective in this scenario is to train a model on the dataset $\mathbf{X}$ by learning one or more self-supervised proxy tasks designed to capture the normality of the data samples. Once the model is trained, the acquired knowledge is utilized to assess the degree of abnormality in new samples, which are then categorized as either normal or anomalous. Specifically, for each new sample $X_{new}$, the trained model is used to calculate its associated anomaly score $\textnormal{AS}(X_{new})$. By this, the higher $\textnormal{AS}(X_{new})$ is, the more likely $X_{new}$ is an anomalous time series.

The studies categorized here, along with their specific characteristics, are presented in Table \ref{tab:global}. This table provides details on the self-supervised task(s) utilized, the metric used for calculating anomaly scores, and whether the method is applied to univariate or multivariate time series data. As in the previous section, the column referring to the anomaly scores outlines different methods for their computation, including evaluating the model's classification performance (including classification losses and misclassifications), assessing residual errors (such as reconstruction and prediction errors), and utilizing metrics associated with the model's latent representations (involving representation similarities and distances). Contrastive losses are also included in this category, as they calculate similarity and distance measures between the latent representations of data segments or samples.

\begin{table}
\begin{center}
  \caption{Summary of self-supervised methods for time series anomaly detection in the global context}
  \label{tab:freq}

  \begin{tabular}{lcccc}
    \toprule
    Paper & Type of PT &Anomaly Score & \multicolumn{2}{c}{Dim}\\
    \cline {4-5}  & & &U & M \\
    \midrule
    Self-predictive & & & & \\
    \hline

    \cite{blazquez2021water} & Class  &  CP & \checkmark &\\
    \cite{kim2021self} & Class &  CP & \checkmark &\\
    \cite{xu2020anomaly} & Class  & CP & &\checkmark\\
    \cite{zhang2021self} & Class  &  CP & &\checkmark\\
    \cite{zheng2022task} & Class  &  RM & &\checkmark\\

    \cite{hayashi2022ocstn} & Rec  &  RE & \checkmark&\\
    
    \cite{wang2022identification} & Rec + Class &  RE & &\checkmark \\
    \cite{zhang2022adaptive} & Rec + Class  &  RE & &\checkmark \\
    \cite{jiao2022timeautoad} & Rec + Class  &  RM & &\checkmark \\
    \cite{bai2023ssdpt} & Rec + Class  &  CP + RE & \checkmark &\\

    \hline
    Contrastive& & & &   \\
    \hline
    \cite{qiu2021neural}   & AC  & RM & & \checkmark\\
    \cite{zhang2022self} & AC  &   RM & & \checkmark \\
    \cite{wang2023deep} & SC + AC &   RM & & \checkmark \\
    
    \hline
    Multi-type&  & & &  \\
    \hline

    \cite{li2022data}  & Class + AC &  RM & \checkmark &\\
    \cite{hojjati2022acoustic}   & Class + AC & RM & \checkmark&\\
    \cite{zeng2023joint}   & Class + Rec + SC &  RE + RM & \checkmark&\\

  \bottomrule
  
\end{tabular}\label{tab:global}

\begin{minipage}{11.7cm}
\small  U: Univariate; M: Multivariate // Class: Classification; Rec: Reconstruction; SC: Sampling Contrast; AC: Augmentation Contrast // Subseq: Subsequence// CP: Classification Performance; RE: Residual Errors; RM: Representation Measures
\end{minipage}
\end{center}
\end{table}

\subsection{Single-type self-supervised learning}

Single-type self-supervised learning methods in this section employ a single type of task, such as self-predictive or contrastive, to identify outlier time series.

\subsubsection{Self-predictive approaches}

Recall from previous sections that self-predictive proxy tasks operate at the data-sample level. In the context of global time series anomaly detection, the models are fed with samples representing complete time series to capture the normality by means of one or more proxy tasks. Once the pretext tasks are learned on the basis of normal samples, the models leverage the acquired knowledge to compute the anomaly score of new samples to be classified as normals or abnormals.

Most methods in this category utilize self-supervised classification-based proxy tasks. In this approach, a set of $K$ transformations is applied to each input sample, resulting in $K$ augmented views per sample in the training set. The model is trained to predict the transformation applied to each input sample for generating its augmented views, thus creating a self-supervised classification with $K$ possible classes (one for each transformation). The augmented views of training samples serve as inputs to the model, while their associated transformations serve as pseudo-labels for the self-supervised classification. 

The main difference among the approaches that employ self-supervised classification for time series anomaly detection relies on the choice of the transformations considered to augment the input samples. In this context, the criteria that must fulfill the transformations are twofold:

\begin{itemize}
    \item \textbf{Break the normality}. The transformations considered need to generate augmented views that disrupt the normal nature of the data, generating `corrupted' versions of the samples that do not follow the normal patterns and characteristics of the data distribution. Note that the selection of a transformation that breaks the normality of the data depends on the context of the problem at hand. Therefore, the augmentations that fulfill this requirement in some problems might not be suitable for performing anomaly detection in other problems.
    \item \textbf{Diversity}. The transformations considered should generate augmented views that share relevant semantic information with the original data. Optimally, the resulting augmented views should produce diverse views of each sample without redundancies on the information they keep.
\end{itemize}

The literature presents various approaches employing manually predefined transformations for self-supervised classification in global time series anomaly detection. For example, \cite{xu2020anomaly} proposes resizing normal training time series at different scales, with each scale representing a distinct class. Other works suggest applying $K$ different affine transformations to input samples to generate augmented views for each considered transformation \cite{blazquez2021water, zhang2021self}. In another approach by \cite{zheng2022task}, they propose to augment normal samples with two transformations: one altering the amplitude and the other modifying the frequency of the time series. The proxy task involves distinguishing between the original samples, the amplitude-changed augmentations, and the frequency-modified views.

Following model training, new samples undergo augmentation with the same transformations used during training, followed by the reapplication of the self-supervised classification task. Subsequently, the model's classification performance concerning the augmented views of the new sample is evaluated to compute its anomaly score. In the approaches of \cite{xu2020anomaly, zhang2021self}, the classification loss utilized during training is employed in the inference phase to measure the model's classification performance and compute the anomaly score of new samples. Conversely, \cite{blazquez2021water} quantifies the number of misclassifications of augmented views of new samples, generating a discrete anomaly score for time series anomaly detection. In contrast, \cite{zheng2022task} utilizes the latent representations of new samples to compute the anomaly score. During inference, latent representations of normal time series in the training set are derived from the model's last hidden layer and represented by a multivariate Gaussian distribution. Subsequently, the anomaly score of new sequences is calculated as the Mahalanobis distance between their latent representations and the Gaussian distribution of normal representations in the model's latent space.

While less prevalent, some studies diverge from utilizing transformations to formulate self-supervised classification tasks. A notable example is the research conducted by \cite{kim2021self}, which opts for leveraging domain-specific knowledge pertaining to the targeted problem: the DCASE2020 challenge \cite{dohi2022description}. Prior to presenting the contribution, we will provide a concise overview of the particulartities of this problem and its configuration. The DCASE2020 challenge revolves around acoustic anomaly detection, specifically identifying irregular sounds in various types of operating machines. Each machine, from which the sounds are recorded, is assigned a unique ID. Thus, each time series in the training and test datasets has an associated ID representing the specific machine from which it was generated \cite{dohi2022description}. Due to this, there are multiple contributions that propose self-supervised approaches for addressing this task. These approaches leverage the "privileged information" inherent in machine IDs to construct self-supervised pretext tasks for addressing the anomaly detection task effectively. In this scenario, the authors of \cite{kim2021self} utilize the machine IDs to form a self-supervised classification task centered on predicting the machine ID associated with each time series in the training set. In the inference phase, as in the previous approaches, anomalous time series are detected by assessing the model's performance in predicting the ID of new time series.

In addition to self-supervised classification, we also introduce the method proposed by \cite{hayashi2022ocstn} within the realm of self-predictive approaches for detecting anomalous sequences. In this study, the authors aim to transform each time series in the training set into a predetermined target signal, which remains fixed as a hyperparameter. To achieve this, they employ a neural network to model the black box function that links the normal time series in the training data with the target signal, thereby capturing the normal distribution of the data. Subsequently, during the inference phase, they repeat the same procedure for new time series, using the prediction error as the anomaly score. We emphasize the resemblance this method shares with self-supervised reconstruction. However, in this instance, the task's objective is not to reconstruct the original time series but to reconstruct a predefined target signal.

As we have seen, there are multiple works that consider only using self-supervised classification for time series anomaly detection. There is only one method that makes use only of self-supervised reconstruction for this aim. However, we find various studies in which self-supervised reconstruction is employed alongside self-supervised classification to enhance the capabillities of the models to detect abnormal time series.   

The works of \cite{wang2022identification, zhang2022adaptive} advocate for a fusion of autoencoder-based unsupervised reconstruction with an auxiliary self-supervised classification. They utilize an autoencoder to reconstruct input time series while simultaneously learning a self-supervised classification task. The transformations considered to perform the self-supervised classification include jittering, reversing and scaling, each resulting in a different class to be distinguished. Additionally, representational memory modules are introduced into the models to mitigate issues stemming from noisy information in the inputs and improving their ability to reconstruct normal samples. Once the tasks are learned concurrently, the anomaly score for new time series is based on the reconstruction errors of the autoencoders in both approaches.

Similarly, the authors in \cite{jiao2022timeautoad} also utilize both self-supervised reconstruction and self-supervised classification. For self-supervised classification, they employ transformations such as shifting and up-scaling segment values, introducing noise, and swapping two random segments. Concurrently, the model adopts an encoder-decoder structure to learn reconstructing the original time series. Additionally, they concatenate the reconstruction error and the latent representation of the original time series and fit them to a Gaussian Mixture Model. These tasks are learned simultaneously, and during inference, the anomaly score of new time series depends on the density of the concatenation of the reconstruction error and the latent representation with respect to the Gaussian Mixture Model. Notably, this study incorporates automated machine learning, employing techniques to automatically select optimal hyperparameters for the machine learning model. It is important to highlight that in this method the anomaly scores are not derived from the proxy tasks during inference but from the Gaussian Mixture Model, which is introduced during model training to better capture the normal characteristics of the data.

Finally, another work proposes a combined approach using a self-supervised reconstruction and a self-supervised classification to solve the DCASE2020 challenge and find abnormal sequences in the global context of time series anomaly detection \cite{bai2023ssdpt}. Given the nature of the problem, they extract the spectrograms from the normal time series sequences. To augment the training data, they mix spectrograms from different machine IDs by creating randomized linear combinations. The self-supervised reconstruction task involves masking parts of these augmented views and then reconstructing them. In addition, the self-supervised classification task aims to predict the weights of the machine IDs associated to the spectrograms used in generating the augmented views. These two proxy tasks are simultaneously learned using a multi-task learning approach. For inference, each new time series serves as input for the model, and the anomaly score is computed by a weighted sum of the reconstruction error of that input and the prediction error for the associated machine ID.

\subsubsection{Contrastive approaches}

These methods utilize one or more contrastive pretext tasks to train the models. The representations learned from these tasks are then utilized to determine the anomaly score of new time series, distinguishing them as normal or abnormal. Specifically, all methods in this category employ augmentation contrast, where transformations are applied to input time series samples to generate pairs for the contrastive task. Positive transformations aim to impart invariance to the model regarding normality-preserving changes, while negative samples are created by applying transformations that disrupt normality or selecting other samples from the dataset. The primary distinction among these approaches lies in the choice of transformations used to augment samples, which depends on the assumptions regarding the normal patterns and characteristics of the data made by the authors for each specific problem.

The proposal of \cite{zhang2022self} introduces a contrastive pre-training framework based on three augmentation-based contrastive tasks for time series representation learning, followed by time series anomaly detection. For each time series, this method extracts a time-based representation and a frequency-based representation by means of two encoders. The first two contrastive tasks are based on time- and frequency-consistency. Specifically, the time-based and the frequency-based representations are augmented by means of transformations applied in the temporal domain (jittering, scaling and time shifts) and in the frequency domain (removing and adding frequency components), respectively. Then, these two tasks take each of the two representations with their respective augmented views as the positive pairs, while considering the augmented views of other sequences in the batch as the negatives to the anchor. The third task focuses on ensuring that the distance between the frequency-based and temporal-based representations of the anchor is greater than the distance between the frequency-based and temporal-based representations and their respective augmented views. These tasks are concurrently learned in a multi-task manner during pre-training. For fine-tuning, a one-class SVM is incorporated at the top of the model to handle the anomaly detection task. The one-class SVM establishes a boundary encapsulating the majority of normal instances in the feature space, classifying sequences falling outside that boundary as anomalies during inference.

The previous approach utilizes augmented views of different samples as negative samples for the contrastive task. However, some approaches, such as COCA \cite{wang2023deep}, do not incorporate negative samples in their contrastive tasks. COCA is a deep contrastive approach specifically designed for time series anomaly detection. This method integrates a contrastive pretext task with SVDD, which has been explained in the previous section. SVDD serves as a popular sampling contrast-based method that does not rely on negative pairs. Nevertheless, like other contrastive methods, it may suffer from a drawback known as `hypersphere collapse', where representations of all normal instances converge to a constant. COCA proposes to combine the SVDD task with an augmentation contrast-based contrastive task. To construct the positive pairs of the anchor sequence, jittering and scaling transformations are used. Concurrently, COCA employs the SVDD loss by additionally incorporating a variance term to prevent hypersphere collapse. During inference, the anomaly score for a new time series is computed based on the distance between the representations of the original input and its augmented view relative to the center of the hypersphere.

The methods we have seen in this section make use of manually predefined transformations to generate the pairs to learn the contrastive tasks. However, there is a line of research on developing automatic augmentations that are not manually defined. In this line, we find NeuTraL AD \cite{qiu2021neural}. In this work, the authors introduce a contrastive framework that augment neural network-based transformations, which can be represented by any parameterized function with gradient-based optimization accessibility. For each time series in the training dataset, they apply a set of learnable transformations to create multiple augmented views for each input. In this case, they implement the transformations by means of feed forward neural networks. Then, they employ what they term the `Deterministic Contrastive Loss' to bring together the representations of the augmented views and the original input while simultaneously pushing away the embeddings of the augmented views away from each other. Note that the augmentation neural network is learned during the training phase as well. During the inference phase, the training loss serves as the anomaly score for effectively identifying anomalous time series.

\subsection{Multi-type approaches}

This section reviews methods that utilize both self-predictive and contrastive proxy tasks for global time series anomaly detection. Concretely, they all aim to tackle the previously outlined DCASE2020 challenge. Within this category, we identify three works that integrate various approaches discussed earlier to more effectively capture the normal patterns within collections of time series.

The works by \cite{li2022data, hojjati2022acoustic} combine self-supervised classification with an augmentation contrast-based proxy task. Considering transformations like pitch shift, time stretch, and time shifting, two random transformations are selected and applied to input sequences to create two augmented views for each sequence. Subsequently, mel-spectrograms are extracted from these augmented views, forming the basis for jointly learning the tasks. The self-supervised classification task involves predicting the transformation applied to the input for generating each of its two augmented views. The contrastive task pairs the augmented views of the same input as positive, while the rest of the samples in the batch serve as negative samples. After model training, the anomaly score for a new time series is determined by computing the Mahalanobis distance between its representation and the representations of normal training instances.

Lastly, \cite{zeng2023joint} presents another method to tackle the DCASE2020 challenge. This work builds upon previous contributions that leverage machine IDs to extract information from acoustic time series for self-supervised representation learning. In this study, the authors perform self-supervised reconstruction by masking original sequences and reconstructing them to pre-train the model. Following this, they concurrently learn a self-supervised classification and a contrastive task. For the self-supervised classification, each input in the training set is encoded and reconstructed, and a binary classifier is employed to distinguish between the original time series and their corresponding reconstructions. To enhance the model's discriminative ability with respect to original and reconstructed samples, a contrastive task is introduced, considering samples with the same ID as positive pairs and their reconstructed counterparts as negative samples. During inference, the model's reconstruction error and the similarity of samples with the same ID are utilized to compute the anomaly score of new sequences.

\section{Available Software}\label{sec:software}

Within this section, we compile the openly accessible software related to the self-supervised time series anomaly detection approaches delineated in the preceding sections. An overview of this software is presented in Table \ref{tab:software}. The organization of the table is structured in accordance with the context in which anomaly detection is performed and the type of anomalies they aim to identify. The method's name contains a hyperlink leading to the URL for access. The summary of the most used datasets for evaluating the performance of self-supervised time series anomaly detectors is presented in Table \ref{tab:datasets} (see Appendix \ref{sec:datasets}).

\begin{table*}
\begin{center}
  \caption{Summary of the publicly available software associated to self-supervised time series anomaly detection methods.}
  \begin{tabular}{lcc}
    \toprule
    Name&Related Research&Anomaly Type\\
    \midrule
    Local anomaly detection & &  \\
    \hline
    \href{https://github.com/Jhryu30/AnomalyBERT}{AnomalyBERT} & \cite{jeong2023anomalybert} & Point \\
    \href{https://github.com/qiumiao30/SLMR}{SMLR} & \cite{miao2022unsupervised} & Point \\
    \href{https://github.com/yuezhihan/TS2Vec}{TS2Vec} & \cite{yue2022TS2Vec} & Point \\
    \href{https://github.com/boschresearch/local_neural_transformations}{LNT} & \cite{schneider2022detecting} & Point\\
    \href{https://github.com/DAMO-DI-ML/KDD2023-DCdetector}{DCdetector} & \cite{yang2023dcdetector} & Point\\
    \href{https://github.com/anhduy0911/CoInception}{CoInception} & \cite{nguyen2023learning} & Point\\
    \href{https://github.com/manigalati/usad}{USAD} & \cite{audibert2020usad} & Point\\
    \href{https://github.com/wxdang/MSCRED}{MSCRED} & \cite{zhang2019auto} & Subsequence\\

    \href{https://github.com/xuhongzuo/couta}{COUTA} & \cite{xu2022calibrated} & Subsequence \\

    \href{https://github.com/KDD-OpenSource/ContrastAD}{ContrastAD} & \cite{li2023contrastive} & Subsequence\\
    \href{https://github.com/datacubeR/DeepAnt}{DeepAnt} & \cite{munir2018deepant} & Point, Subsequence\\
    \href{https://github.com/WaRTEm-AD/UnivariateAnomalydetection}{WaRTEm-AD} & \cite{abilasha2022warping} & Point, Subsequence\\
    \hline
    Global anomaly detection & &  \\
    \hline
    \href{https://github.com/ToshiHayashi/OCSTN}{OCSTN}  & \cite{hayashi2022ocstn}  & Sequence\\
    \href{https://github.com/JishengBai/SSDPT}{SSDPT} & \cite{bai2023ssdpt} & Sequence\\
    \href{https://github.com/boschresearch/NeuTraL-AD}{NeuTraL AD} & \cite{qiu2021neural} & Sequence \\
    \href{https://github.com/Armanfard-Lab/AADCL}{AADCL} & \cite{hojjati2022acoustic} & Sequence\\
    \href{https://github.com/ironing/eeg-ad}{-}& \cite{zheng2022task} & Sequence\\
    \href{https://github.com/ruiking04/COCA}{COCA} & \cite{wang2023deep} & Sequence \\
    \href{https://github.com/zhangyuxin621/AMSL}{AMSL} & \cite{zhang2022adaptive} & Sequence \\
    \href{https://github.com/mims-harvard/TFC-pretraining}{TF-C} & \cite{zhang2022self} & Sequence \\

  \bottomrule
\end{tabular}\label{tab:software} 
\end{center}
\end{table*}

\section{General Discussion and Conclusions}\label{sec:conclusions}

Time series anomaly detection is an expanding area of research due to its relevance in several application domains. With the emergence of novel machine learning techniques, numerous researchers have incorporated self-supervised learning into their methodologies to overcome the limitations of traditional unsupervised methods and improve the efficacy of anomaly detection algorithms. In this paper, we have explored the current literature of self-supervised methods designed for time series anomaly detection. In addition, we have introduced a taxonomy that categorizes these approaches based on the context in which they solve the anomaly detection task, and the type of pretext tasks considered. This final section offers general remarks on the examined works and presents specific conclusions organized according to the primary axes of the proposed taxonomy. Together with the conclusiones, we provide some insights about possible future research directions in the field.

To begin, all the works we have explored are rooted on a one-class classification perspective, in which the training set is assumed to comprise only normal samples. The main advantage of these methods is that they do not rely on the use of human annotations about normal and abnormal samples for model training. However, this is a heavy assumption specially in real problems, as there might exist samples or data parts that are anomalous in our training set. Thus, as unsupervised models tend to be sensitive to training set contamination due to the presence of unrecognized anomalous patterns, there is room for the design of methods that are robust to the existence of abnormalities in the data used for model training. Consequently, as a potential direction for future research, exploring whether self-supervised approaches demonstrate reduced sensitivity to the contamination of the normal training set would be interesting.

Concerning the inference phase of anomaly detection models, one of the key contributions of the articles discussed in this paper lies in introducing novel methods for computing anomaly scores for new data leveraging self-supervised learning. These scores measure the degree of abnormality in new samples, aligning with the ultimate goal in anomaly detection. Nevertheless, numerous works refrain from explicitly defining a threshold value for identifying anomalies due to the absence of a precise and clear methodology. Hence, there is potential value in exploring new research endeavors in this aspect.

As a final general conclusion, we observe that the proposed methods exhibit diverse and intriguing properties, showcasing a variety of approaches to time series anomaly detection. Nevertheless, as outlined in Section \ref{sec:software}, only a limited number of works have shared the source code of their proposed methods for public access. To encourage ongoing progress in this field, it is important to emphasize the significance of releasing the source code of new methods. This practice facilitates other researchers in comprehending these approaches thoroughly and refining their own proposals. Moreover, it would be valuable to create new datasets for evaluating newly proposed approaches. Since many contributions undergo testing on the same datasets, there is a risk of creating an inaccurate perception of progress in the advancement of techniques for time series anomaly detection \cite{wu2021current}.

We shift our focus fto each axis individually, beginning with the first one. The approaches we have analyzed address the challenge of anomaly detection in two contexts: (i) identifying abnormal points and subsequences at the local level of individual time series, and ii) detecting anomalous sequences within the global context of a dataset composed of a number of time series. Most of the analyzed methods fall into the former category, where long time series are typically segmented into subsequences to analyze local data dynamics. While numerous works aim to identify abnormal points in the local context, there are a few approaches focused on detecting anomalous subsequences. Both types of approaches aim to capture the temporal dynamics and characteristics of individual time series. Thus, by adjusting the computation of anomaly scores from points to subsequences, many of the methods within this framework that have been examined may effectively identify both abnormal points and subsequences in the local context of time series. It would be interesting to test if the approaches falling under this category are capable of accurately detecting both point and subsequence outliers by means of the previously mentioned adjustment.

In the realm of local time series anomaly detection, there exists a particular task based on the identification of point and subsequence anomalies in streaming scenarios. Detecting anomalies in real-time or streaming settings involves recognizing abnormal patterns or events in an ongoing data flow. In this context, data points arrive sequentially, emphasizing the need to promptly detect anomalies as they unfold. This is particularly critical in applications where timely identification of unusual behavior is crucial. In the literature, only a limited number of works address the local anomaly detection task in streaming scenarios \cite{jiang2017wind, chen2020anomaly, yue2022TS2Vec}. In this sense, as a future research direction, the proposal of self-supervised methods that address time series anomaly detection in streaming contexts is crucial for improving the robustness and reliability of time series anomaly detection in real systems systems.

In the context of global time series anomaly detection, these approaches strive to identify overarching patterns and characteristics representing the normality of data across datasets comprising diverse time series. Even if there are not many, there are global anomaly detection-based approaches that divide the original sequence into subsequences and treat them as complete sequences within the dataset. This strategy, presented by works such as \cite{meng2019spacecraft, huang2022efficient, tran2022self}, suggests that many global time series anomaly detection methods can be adapted for local anomaly detection by segmenting the original time series into subsequences and treating each subsequence as a distinct time series. 

Concerning the second axis of the taxonomy, which pertains to the types of pretext tasks in the self-supervised setting, a notable prevalence of single-type approaches is observed in comparison to multi-type approaches, particularly methods based on self-predictive tasks rather than on contrastive tasks. Let us analyze the fundamental attributes of the proxy tasks inherent in these two categories of self-supervised learning.

In reference to self-predictive learning, the explored approaches usually rely on the use of transformations for constructing proxy tasks. Within this field, we specifically remark the utilization of self-supervised classification and self-supervised reconstruction tasks within this domain. Based on the analyzed works, it is notable that self-supervised classification tasks (alone or in combination with self-supervised reconstruction), are predominantly utilized in global anomaly detection to capture general patterns that describe the normality of collections of samples. In contrast, self-supervised reconstruction and forecasting are more commonly employed to capture local patterns and characteristics for identifying abnormal points and subsequences.

In the context of self-supervised classification, we see that the proposed transformations are typically chosen to satisfy the following criteria: i) disrupting the normality of the data to capture anomalous behavior when subjected to transformations and performing the self-supervised classification task, and ii) ensuring that the resulting augmented views are as diverse as possible \cite{qiu2021neural}. Many methods in this field advocate for manually designed transformations to fulfill these two conditions depending on the problems they aim to tackle \cite{huang2022efficient, tran2022self, jeong2023anomalybert, blazquez2021water, zhang2021self, zheng2022task, xu2020anomaly}. The most popular choices include resizing and scaling, applying affine transformations or altering the frequency of the time series. The performance of these methods heavily relies on their chosen transformation accurately fitting the data, so a transformation effective for one problem may not be suitable for another. Motivated by the proposal of NeUtraL AD \cite{qiu2021neural}, there is a line of research based on generating augmented views of samples for contrastive tasks on the basis of neural network-based transformations, which are jointly learned with the remaining modules of the model proposed \cite{chen2023time, schneider2022detecting, zheng2023unsafe}. Inspired by these works, a possible approach could be to employ neural networks to learn automatic transformations that meet the previous two conditions and employ them in self-supervised classification-based time series anomaly detection. This could lead to self-supervised classification-based methods applicable to many different problems without depending on the manual design of effective transformations applicable to all of them.

Switching to the domain of self-supervised reconstruction and forecasting, the most widely utilized transformations are masking \cite{fu2022mad, liu2021deepfib, pan2022duma, meng2019spacecraft, sakurada2014anomaly, jiang2017wind} and jittering \cite{marchi2015novel, sakurada2014anomaly}. Even if these two methods for generating augmented views show good results in terms of anomaly detection, it would be interesting to propose more transformations for self-supervised reconstruction. Moreover, it would also be possible to combine different transformations to generate augmented views as this could lead to methods that are more robust to perturbations on the data and achieve a better performance.

Concerning contrastive tasks, methods favoring augmentation contrast are more prevalent than those advocating for sampling contrast. Augmentation contrast typically involves manually predefined transformations to create invariance to positive transformations while differing from negative ones \cite{li2023contrastive, chambaret2022stochastic, carmona2021neural}. In addition, as mentioned before, recent advancements in deep learning have led to the adoption of neural transformations in the contrastive setting \cite{chen2023time, schneider2022detecting, zheng2023unsafe}. Sampling contrast is primarily adopted in methods focused on local anomaly detection, with the most common approach being CPC \cite{pranavan2022contrastive, kang2023tictok}, which shares some similarities with self-supervised forecasting but belongs to the contrastive setting. Additionally, certain works leverage "privileged information" specific to the problem, such as machine IDs in the DCASE2020 problem \cite{zeng2023joint}, to construct contrastive pairs for learning these tasks. For computing anomaly scores for new samples, most approaches utilize measures associated with the latent representations extracted from the model. 

Finally, only a few works opt for multi-type self-supervised learning in time series anomaly detection. These approaches employ multiple proxy tasks, including both self-predictive and contrastive tasks, during the training of the models. After training, some tasks may be used, while others may be discarded as they are meant to be auxiliary for learning better representations during the training of the models. In the literature, the most common combination involves using autoencoders with other pretext tasks \cite{abilasha2022warping, yang2022unsupervised, zhou2022contrastive, zhang2022timegrid}.

Following the above, multi-type approaches have the potential to harness the benefits of both self-predictive learning, which focuses on acquiring robust representations at the data sample level, and contrastive learning, which excels at capturing patterns related to relationships between different data parts and samples. Therefore, further exploration of methods combining self-predictive and contrastive tasks is merited, as they can extract both high and low-level features crucial for capturing data normality in anomaly detection tasks.

\section * {Acknowledgements}
Authors thank financial support of Ministerio de Economía, Industria y Competitividad (MINECO) of the Spanish Central Government [PID2019-104933GB-10/AEI/10.13039/501100011033 and PID2022-137442NB-I00], and Departamento de Industria of the Basque Government [IT1504-22 and ELKARTEK Programme].  
A. S. F. thanks financial support of Departamento de Educación of the Basque Government under the grant PRE\_2022\_1\_0103. 
JA. L. thanks financial support of the Basque Government through the BERC 2022-2025 program and
by the Ministry of Science and Innovation: BCAM Severo Ochoa accreditation CEX2021-001142-S / MICIN / AEI / 10.13039/501100011033.

\bibliographystyle{ACM-Reference-Format}
\bibliography{libs}

\appendix

\section{Methodology}\label{sec:methodology}

To conduct this survey and ensure its reproducibility, we have implemented a systematic methodology consisting of four key processes: Database Selection, Literature Search, Selection of Relevant Studies, and Analysis of the Studies. Each of these stages contributes to the comprehensive investigation of the research topic under consideration.

\medskip

\textit{Database Selection}. For this survey, we have chosen several scientific research databases to gather relevant publications. The selected databases are: ACM Digital, IEEE Explore Digital Library, ScienceDirect, SpringerLink, DBLP, Google Scholar, and Semantic Scholar.

\medskip

\textit{Literature Search}. To find articles that are relevant to the subject of this survey, we employ a literature search strategy using different queries. These queries are created by combining various strings that represent the methodology, task, and type of data considered. Specifically, we generate all possible queries resulting from the concatenation of the following strings:

\medskip

\begin{center}

    (Self-Supervised Learning \textit{OR} Contrastive Learning) \\
    \textit{AND}\\
    (Anomaly Detection \textit{OR} Outlier Detection) \\
    \textit{AND} \\
    "Time Series Data"
\end{center}

\medskip

Additionally, we also examine the papers referenced by the selected studies and those that cite them, as long as they are relevant to the subject of our survey.

\medskip

\textit{Selection of Relevant Studies}. In our analysis, we consider contributions that explicitly state the use of self-supervised learning for detecting anomalies in time series data in their titles or abstracts. We include these studies in our survey to examine their findings and approaches.

\medskip

\textit{Analysis of the Studies}. Upon identifying and selecting the relevant contributions for analysis, we proceed to extract their primary characteristics in order to construct a comprehensive taxonomy for classifying these studies. In addition to this, we conduct a detailed examination of other properties that characterize each of the studies.

\section{Dataset list}\label{sec:datasets}

Many methods use well-known time series datasets to evaluate models in anomaly detection tasks. Table \ref{tab:datasets} presents the main characteristics of the most commonly used datasets for evaluating self-supervised time series anomaly detection approaches. These properties include the dataset's name, associated scientific research, types of outliers it contains, the number of dimensions considered in the time series. The dataset's name contains a hyperlink leading to the URL for access.

\begin{table*}
\begin{center}
  \caption{Most popular datasets in time series anomaly detection together with their properties}
  \begin{tabular}{lccc}
    \toprule
    Name&Related research&Anomaly Type& Dim\\
    \midrule
    \href{https://webscope.sandbox.yahoo.com/catalog.php?datatype=s&did=70}{Yahoo-TSA} &  \cite{laptev2015benchmark} & Point & $1$ \\
    \href{https://dataverse.harvard.edu/dataset.xhtml?persistentId=doi:10.7910/DVN/6C3JR1}{Tennessee Eastman Process}&  \cite{TEEEEP} & Point & $52$\\
    \href{https://nsidc.org/data/smap/data}{SMAP} & \cite{hundman2018detecting}  & Point, Subsequence & $25$ \\
    \href{https://github.com/NetManAIOps/OmniAnomaly/tree/master/ServerMachineDataset}{SMD} &  \cite{su2019robust} & Point, Subsequence & $38$\\
    \href{https://itrust.sutd.edu.sg/itrust-labs_datasets/dataset_info/}{SWaT} & \cite{mathur2016swat} & Point, Subsequence & $51$\\
    \href{https://pds-atmospheres.nmsu.edu/data_and_services/atmospheres_data/Mars/Mars.html}{MSL} & \cite{hundman2018detecting}  & Point, Subsequence & $55$\\
    \href{https://itrust.sutd.edu.sg/itrust-labs_datasets/dataset_info/}{WaDi} & \cite{ahmed2017wadi} & Point, Subsequence & $123$ \\
    \href{https://zenodo.org/record/4740355}{MIMII}& \cite{purohit2019mimii}&Sequence & $1$ \\
    \href{https://zenodo.org/records/3351307#.XT-JZ-j7QdU}{ToyADMOS} & \cite{koizumi2019toyadmos} & Sequence & $1$ \\
    \href{https://www.kaggle.com/datasets/nxthuan512/upennmayoeegdatadog1}{UPenn and Mayo Clinic's Seizure} & \cite{UPMC} & Sequence & $16$  \\
    \href{https://www.cs.ucr.edu/~eamonn/time_series_data_2018/}{UCR Time Series Classification Archive$^*$} &  \cite{dau2019ucr} & Sequence & $-$ \\

    \href{https://wu.renjie.im/research/anomaly-benchmarks-are-flawed/}{UCR Anomaly Archive }&  \cite{wu2021current} & Point, Subsequence, Sequence & $-$ \\
    
  \bottomrule
\end{tabular}\label{tab:datasets} 
\begin{minipage}{16.9cm}
\small $^*$Unbalanced classification datasets are often employed for anomaly detection by designating the majority class as the normal category and categorizing the remaining classes as anomalous.
\end{minipage}
\end{center}
\end{table*}

\end{document}